\documentclass{article}

\PassOptionsToPackage{numbers,sort&compress}{natbib}

\usepackage[preprint]{neurips_2025}

\usepackage[utf8]{inputenc} 
\usepackage[T1]{fontenc}    
\usepackage{hyperref}       
\usepackage{url}            
\usepackage{booktabs}       
\usepackage{amsfonts}       
\usepackage{nicefrac}       
\usepackage{microtype}      
\usepackage{xcolor}         

\usepackage{amsmath,amssymb}
\usepackage{algorithmic}
\usepackage{graphicx}
\usepackage{multirow}
\usepackage{enumitem}
\usepackage{subcaption}

\title{IsoNet: Causal Analysis of Multimodal Transformers for Neuromuscular Gesture Classification}

\author{%
  Eion Tyacke\\
  Dept. of Electrical and Computer Engineering\\
  New York University\\
  Brooklyn, NY 11201 \\
  \texttt{et1799@nyu.edu}
  \And
  Kunal Gupta \\
  Dept. of Electrical and Computer Engineering\\
  New York University\\
  Brooklyn, NY 11201 \\
  \texttt{kg3163@nyu.edu}
  \And
  Jay Patel \\
  Dept. of Electrical and Computer Engineering\\
  New York University\\
  Brooklyn, NY 11201 \\
  \texttt{jp5207@nyu.edu}
  \And
  Rui Li \\
  Tandon School of Engineering\\
  New York University\\
  Brooklyn, NY 11201 \\
  \texttt{rui.li@nyu.edu}
}

\begin{document}

\maketitle

\begin{abstract}
Hand gestures are a primary output of the human motor system, yet the decoding of their neuromuscular signatures remains a bottleneck for basic neuroscience and assistive technologies such as prosthetics. Traditional human‑machine interface pipelines rely on a single biosignal modality, but multimodal fusion can exploit complementary information from sensors. We systematically compare linear and attention‑based fusion strategies across three architectures: a Multimodal MLP, a Multimodal Transformer, and a Hierarchical Transformer, evaluating performance on scenarios with unimodal and multimodal inputs. Experiments use two publicly available datasets: NinaPro DB2 (sEMG and accelerometer) and HD‑sEMG 65‑Gesture (high‑density sEMG and force). Across both datasets, the Hierarchical Transformer with attention‑based fusion consistently achieved the highest accuracy, surpassing the multimodal and best single‑modality linear‑fusion MLP baseline by over 10\% on NinaPro DB2 and 3.7\% on HD‑sEMG. To investigate how modalities interact, we introduce an Isolation Network that selectively silences unimodal or cross-modal attention pathways, quantifying each group of token interactions' contribution to downstream decisions. Ablations reveal that cross‑modal interactions contribute approximately 30\% of the decision signal across transformer layers, highlighting the importance of attention‑driven fusion in harnessing complementary modality information. Together, these findings reveal when and how multimodal fusion would enhance biosignal classification and also provides mechanistic insights of human muscle activities. The study would be beneficial in the design of sensor arrays for neurorobotic systems.
\end{abstract}

\section{Introduction}

The integration of biosignals with body movement data represents a cutting-edge approach to enhancing the motor control of prosthetic devices. For decades, prosthetic limbs were designed to restore basic functionality, but they often lacked the fine motor control and sensory feedback that is vital for natural limb movement. The goal is to replicate the complex dexterity and functionality of a natural hand with inputs of biosignals, such as surface electromyography (sEMG), electroencephalography (EEG), and neural signals from the brain.

sEMG and finger force biosignals have gained traction across immersive and assistive technologies: in virtual reality, they provide nuanced information for fine object manipulation and grip intensity \cite{bionic_artificial_hands_review,vr_ar_rehabilitation_emg_review}, and in augmented reality, their combination enables seamless translation of physical gestures into intuitive digital actions \cite{force_aware_vr_ar_emg}. Exoskeleton systems have likewise integrated  sEMG with finger force to align user intent and mechanical assistance during grasping \cite{force_guided_grasping_semg}. Beyond AR/VR applications, the fusion of electrical and mechanical 
insights from sEMG and mechanomyography (MMG) sensors have demonstrated improved gesture recognition for hand prosthetic control \cite{ultra_sensitive_emg_fmg_sensor,control_robotic_prosthetic_hands}.

sEMG captures neural drive but is degraded by noise and muscle-to-muscle crosstalk \cite{semg_signal_quality_review}; MMG is immune to electrical interference yet suffers from motion artifacts and low SNR \cite{wearable_interfaces_gesture_recognition}; finger-force sensors report output force but provide no direct view of neuromuscular activation \cite{prosthetic_hand_control_review}. Combining these channels in a multimodal pipeline offsets the weaknesses of each modality and yields a more complete representation of muscle function.

Multimodal machine learning seeks joint representations for heterogeneous data streams; fusion may occur at raw-input, embedding, or decision levels \cite{foundations_in_multimodal,representation_learning_survey}. Late fusion with transformers is attractive because self- and cross-attention capture long-range dependencies and have proved effective in tasks as diverse as 3-D scene understanding \cite{end-to-end_autonomous_driving}. 

Although cross-attention can align modalities, several studies report that self-attention is just as effective \cite{is_cross-attention_preferable,recurrence_vs_attention_for_affect_recognition}. To clarify the picture for biosignals, we compare three architectures that share a similar parameter budget on \textbf{NinaPro DB2} (sEMG + ACC) and \textbf{HD-sEMG 65-Gesture} (HD-sEMG + force): a linear-fusion MLP that concatenates modality-specific linear embeddings before classification; a dual-encoder Transformer that performs self-attention within each modality head and then concatenates the resulting token streams before classification; and a Hierarchical Transformer that adds additional transformer layers over the merged token sequence before classification.

Since the importance of each modality can vary between datasets, for example, dynamic modality gating in text–video sentiment \cite{crossattention_not_enough} and factorized representations in CMU-MOSI \cite{factorized_multimodal_representations}, raw precision alone does not provide a complete measure of performance. A fair assessment must also quantify how much each modality, and their interactions, contribute to the final decision.

Mechanistic interpretability offers that granularity by tracing functional components \cite{mech_interp_a_review}. We introduce the Isolation Network (IsoNet), which enables modality-edge masking via reusable callbacks that ablate cross-modal or unimodal attention edges without altering token content in transformer-based models. Unlike the patch-in causal tracing used for BLIP vision-language models \cite{towards_vl_mech_interp}, our masks operate directly on attention logits, do not need a noise model, and execute in a single forward pass, making it suitable for real-time inference on embedded prosthetic hardware.

Our key contributions are as follows:

\newcommand{\contrib}[1]{%
  \par\hangindent=1.2cm\hangafter=1%
  \noindent\ #1\par\vspace{2pt}}

  \contrib{ \hspace{0.4cm}•  \textbf{Contribution 1}: We evaluate linear concatenation, single‐stage Transformers, and a hierarchical Transformer on identical biosignal tasks, using IsoNet’s causal masking to measure how impactful uni- and cross-modal attention paths are on accuracy.}

  \contrib{ \hspace{0.4cm}•  \textbf{Contribution 2}: Achieved top gesture classification accuracy using only the first 0.5 s of transient muscle activity, ideal for  real-time applications, with the Hierarchical Transformer reaching 97\% on 40 gestures in NinaPro DB2 and 97\% on 65 gestures in HD-sEMG.}

  \contrib{ \hspace{0.4cm}•   \textbf{Contribution 3}: IsoNet ablations on NinaPro DB2 Exercise B reveal 30.4\% of the predictive signal flows through mid-to-penultimate cross-modal heads, while the first and final layers have limited contribution. This is an important sensor and model design reference.}

These results show that modality-aware, attention-centric architectures not only boost accuracy but also deliver transparent, layer-level insights essential for prosthetic and neurorobotic deployment.

\section{Materials and Methods}

\subsection{The Utilized Dataset}\label{A}

Two datasets were used in this study: NinaPro DB2 and HD-sEMG 65 Gesture. Both datasets provide multimodal data, primarily containing sEMG measurements. In addition, NinaPro records kinematic data and HD-sEMG records force data. In multi-modal applications, NinaPro is suited for prosthetic control while HD-sEMG is suited for AR/VR integrations.

The HD-sEMG 65 Gesture dataset records sEMG using two 8×8 electrode grids placed on the flexors and extensors of the forearm \cite{HDsEMG_dataset}. Force readings were acquired with an isometric force device using nine strain gauges (two for the thumb, three for the wrist, and four for the flexion–extension forces of digits two through five). Data was collected from 20 subjects (14 men and 6 women) performing 65 gestures, categorized as 36 one degree-of-freedom, 25 two degree-of-freedom, and 5 complex gestures. The sEMG signals were recorded at 2048 Hz and the force data at 200 Hz. Preprocessing of the signals involved applying a band-pass filter from 10 Hz to 900 Hz, removing 50 Hz power line interference with a third-order Butterworth filter (4 Hz bandwidth), and normalizing force measurements within a 0–5 V range (with a neutral value of 2.5 V).

The NinaPro DB2 dataset was recorded using eight electrodes evenly distributed around the forearm, with additional electrodes placed on the tricep, bicep, forearm extensor, and forearm flexor \cite{Ninapro_dataset}. This configuration allowed for the simultaneous collection of sEMG and three-dimensional kinematic data (x, y, and z acceleration). Recordings were obtained from 40 subjects (29 men and 11 women) performing 49 hand gestures. The gestures include a subset of 17 basic finger movements and 23 grasping/functional movements. The sEMG signals were sampled at 2 kHz, while the accelerometer data were recorded at 148 Hz and subsequently upsampled. Frequency shielding for both 50 Hz and 60 Hz was applied to reduce interference.

\subsection{Data Preprocessing}\label{B} 

Data preprocessing involved specific gesture selection, filtering, and segmentation for both datasets. For NinaPro DB2, only the first 40 gestures were used, excluding the last 9 force-related gestures. Training data came from repetitions 1, 3, 4, and 6, while repetitions 2 and 5 were used for testing. For HDsEMG, repetitions 1, 3, and 4 were used for training, and repetitions 2 and 5 for testing. Both datasets underwent a fourth-order Butterworth band-pass filter between 10 and 500 Hz to isolate electrical muscle activity and a low-pass filter of 90 Hz for kinematic and force data. The data was z-score normalized and rectified, and only the first 0.5 seconds were used to focus on the transient phase, better reflecting real-time response conditions. 

To ensure a more stable training setup for the mechanistic interpretability experiments, we used only the Ninapro DB2 Exercise B, gestures 1 to 17, and limited the data to the steady-state phase of each repetition, specifically 1 to 2 seconds.

\subsection{Causal Analysis}\label{sec:causal}

We estimate the functional importance of specific attention pathways through causal intervention of the attention mechanism with \emph{selective masking} at inference time. Let $A^{\ell}\!\in\!\mathbb{R}^{H\times T\times T}$ be the raw attention-logit tensor (query–key dot products) at layer~$\ell$ ($H$ heads, $T$ tokens). Given a target edge set $\mathcal{S}\subseteq A^{\ell}$, we overwrite its logits with $-\infty$:

\[
A^{\ell}_{h,i,j}\;\leftarrow\;
\begin{cases}
-\infty & \text{if } (h,i,j)\in\mathcal{S}\\
A^{\ell}_{h,i,j} & \text{otherwise}.
\end{cases}
\]

The subsequent soft-max redistributes probability mass over the unmasked entries, nullifying information flow through~$\mathcal{S}$ while keeping the layer’s stochastic semantics. All model weights remain frozen; only the forward pass is altered.

\textbf{Masking categories.}  
We investigate three disjoint edge sets:  
\textit{Unimodal}, query–key pairs from the same modality; \textit{Cross-modal}, pairs from different modalities; and \textit{Layer ranges}, cumulative masks from the first layer up to~$\ell$ or from the final layer back to~$\ell$, each applied separately to uni- and cross-modal edges.

\textbf{Evaluation.}  
For every combination of layers, masking mode, and masking type, we record the averaged subject accuracy and compute
\[
\Delta_{\ell,\mathcal{S}}^{\%}=100\frac{\mathrm{Acc}_{\text{masked}}-\mathrm{Acc}_{\text{baseline}}}{\mathrm{Acc}_{\text{baseline}}}
\]
to assess the effect of removing the specified edge sets.

\subsection{Statistical Analysis}\label{sec:stats}

All hypothesis tests compare each ablated condition with a baseline for the same set of subjects.   Because normality cannot be assumed, we use the two-sided
Mann–Whitney $U$ test. For the causal-masking study there are ten independent comparisons
($3$ mask modes $\times$ $2$ mask types $\times$ $5$ layers), so each raw $p$ value is multiplied by 10 (Bonferroni correction) and capped at 1.  
The same procedure is applied to other groups of experiments when the number of comparisons differs; the correction factor is always the group-specific
count of the null hypothesis tests.

We define the significance symbols for the corrected $p$-value ranges as

\begin{center}
\begin{tabular}{ccccc}
\toprule
$p_\text{corr}$ & $<\!0.0001$ & $<\!0.001$ & $<\!0.01$ & $<\!0.05$ \\
\midrule
Symbol & \texttt{****} & \texttt{***} & \texttt{**} & \texttt{*} \\
\bottomrule
\end{tabular}
\end{center}

Values with $p_\text{corr}\ge 0.05$ are marked $ns$ (not significant).

\section{Model Structure}
\begin{figure*}[h]
    \centering
    \includegraphics[width=1\linewidth]{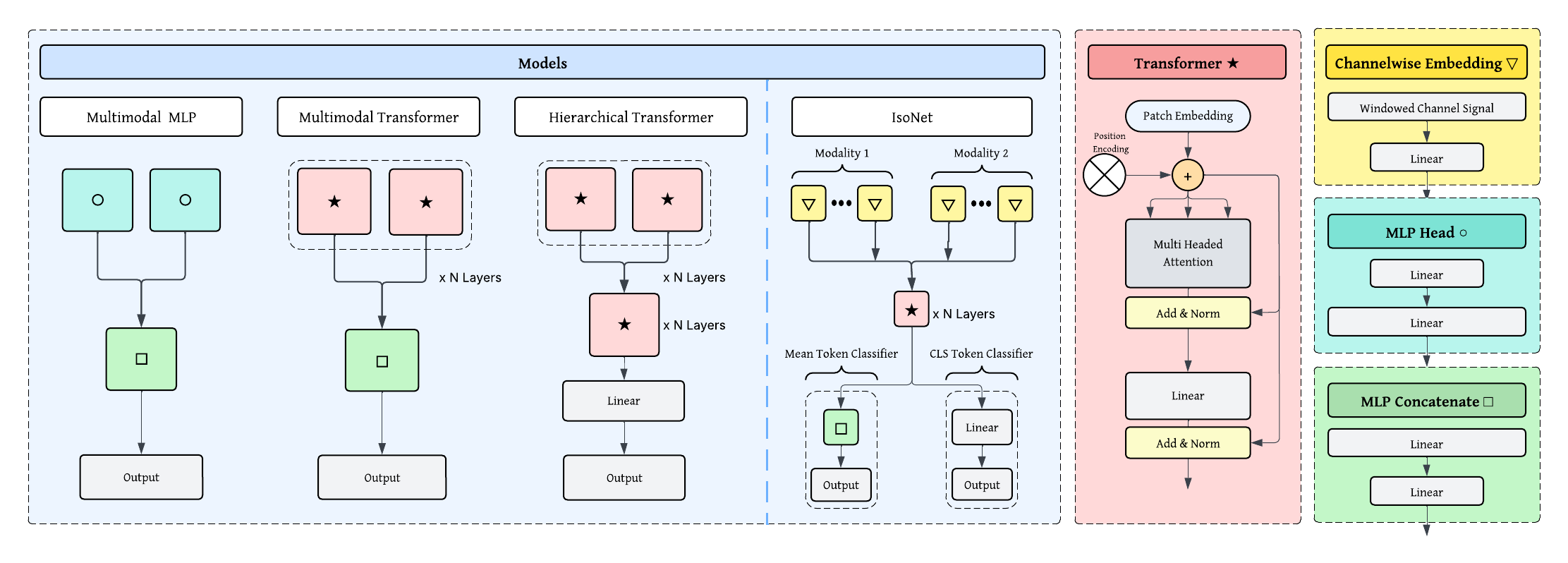}
    \caption{Three multimodal classification architectures are presented, from the left, showing increasing use of transformers to capture cross-modal interactions: (i) linear fusion via MLP, (ii) joint transformer blocks, and (iii) hierarchical transformers, followed by IsoNet, a separate model for mechanistic interpretability. IsoNet encodes windowed channel-wise tokens without early mixing and processes them with shared transformer layers. It outputs dual logits, combined during training via an annealing schedule. Colors and symbols are reused across panels for reference.}
    \label{fig:three_models}
\end{figure*}

Three models were evaluated for multimodal gesture classification using the Ninapro and HD-sEMG datasets: a multimodal multilayer perceptron (MLP), a multimodal transformer, and a hierarchical transformer. These models can be seen in Fig. \ref{fig:three_models})

The multimodal MLP employs dual independent heads (one per modality), each containing two linear layers with ReLU activation. For Ninapro's sparse sEMG and accelerometer data, each head layer uses an input and output embedding size of 200-D. The head outputs concatenate into a shared linear-ReLU layer before final classification.

Both transformer architectures process modal inputs by segmenting data into non-overlapping temporal patches (tubelets) of 40 samples. A classification token is appended to the embedded sequence and positional encoding is added. Each transformer encoder layer integrates multi-head self-attention (modeling token relationships) and a GELU-activated feedforward network, both followed by residual connections, dropout, and layer normalization.

The multimodal transformer processes modalities separately through dedicated transformer modules, concatenating outputs via a linear layer. For Ninapro, the optimal configuration used 512-D embeddings, 8 attention heads, 5 stacked layers, and 128-D feedforward/integration dimensions.

The hierarchical transformer adopts a similar structure but replaces linear fusion with an additional transformer stage to process concatenated head outputs, forming a two-level hierarchy. These architectures progressively substitute linear layers with transformers: the multimodal transformer replaces modality-specific linear processing with dedicated transformers, while the hierarchical transformer further replaces the final linear fusion step with an additional transformer stage to integrate modalities.

The Isolation Network (IsoNet) first maps each channel to its own token via dedicated linear embeddings, providing semantic grounding and preventing any early mixing across channels. Tokens for all channels are concatenated, a learnable CLS token is prepended, and the sequence is processed by the same transformer backbone used in our other multimodal experiments (Fig. \ref{fig:three_models}). IsoNet yields two logits, one from the CLS token and one from the standard mean of non-CLS tokens, trained with an annealed loss
\[
  \mathcal{L}(t) = \lambda(t)\,\mathcal{L}_{\mathrm{CLS}}
               + \bigl[1-\lambda(t)\bigr]\,\mathcal{L}_{\mathrm{avg}},\quad
  \lambda(t)=\min(1,\,t/T_{\mathrm{anneal}}),
\]
so that by epoch 750 only the CLS head drives learning (2,000 epochs total). The average head kicks off training and stabilizes features, while the CLS head provides a clean causal bottleneck for interpretability. Unless otherwise noted, all models were trained using AdamW with a learning rate of 4e-5.

\section{Experiment and Results}

The first experiment evaluates whether multimodal models (trained on combined sEMG and accelerometer [ACC] data) outperform unimodal models (sEMG or ACC alone). While hardware simplification favors single-modality sensor arrays, such designs risk redundancy with diminishing returns. Combining complementary modalities may instead enhance robustness by leveraging cross-modal interactions to mitigate data gaps.

To ensure fair comparison, all models received identical input dimensions and training durations. Data were sourced from the NinaPro armband, which acquires biosignals via 12 electrodes, each providing one sEMG channel and three-axis ACC. Accelerometer magnitudes (computed as $\sqrt{x^2+y^2+z^2}$) yielded one ACC reading per electrode. Electrodes were randomly partitioned into two groups of six across 40 trials, with each trial training three multi-headed neural networks for 200 epochs: (1) sEMG-sEMG (both inputs), (2) ACC-ACC, and (3) sEMG-ACC (multimodal).

\begin{figure}[h]
    \centering
    \includegraphics[width=0.5\linewidth]{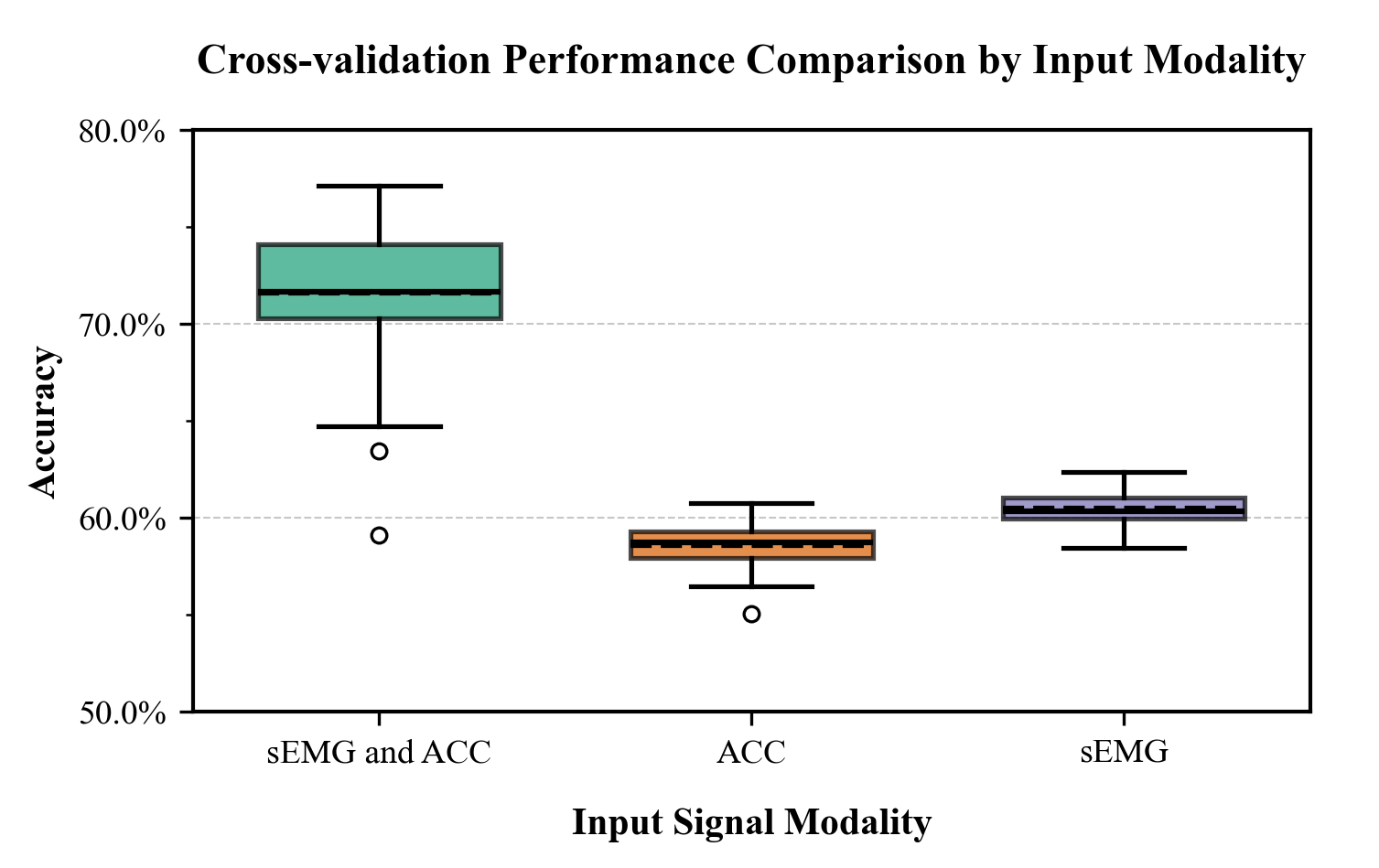}
    \caption{Comparison of three signal types using a multimodal MLP trained on equal amounts of unique data. Each setting uses 12 unique channels: sEMG+ACC (6 sEMG + 6 ACC), sEMG-only (12 sEMG), and ACC-only (12 ACC), repeated over 40 trials. }
    \label{fig:equal_dat_eval}
    \vspace{-0.5cm}
\end{figure}

Figure \ref{fig:equal_dat_eval} shows the multimodal model achieved significantly higher mean accuracy (71.6\%) versus the top unimodal model (60.5\%). sEMG outperformed ACC by 1.9 percentage points (60.5\% vs. 58.6\%), confirming EMG’s stronger discriminative capacity. The ACC model exhibited 1.2 times higher variability (IQR: 57.9–59.3\% vs. 59.9–61.1\% for sEMG), suggesting noise sensitivity. The multimodal model achieved the highest median accuracy (71.7\%) and the highest overall accuracy (77.1\%), demonstrating both consistent performance and the potential for exceptional results. This suggests that sensor fusion can enhance robustness and mitigate the effects of modality-specific noise.

In the second experiment, the three proposed architectures were evaluated on three input modalities (sEMG and ACC, ACC only, and sEMG only) across all 40 subjects trained for 1200 epochs. Each model was trained using only the first 0.5 s of the biosignal (the transient response), making this a substantially more challenging classification task. 

\begin{figure}[h]
\centering
\captionsetup{justification=centering}
\begin{subfigure}[t]{0.48\linewidth}
    \centering
    \includegraphics[width=\linewidth]{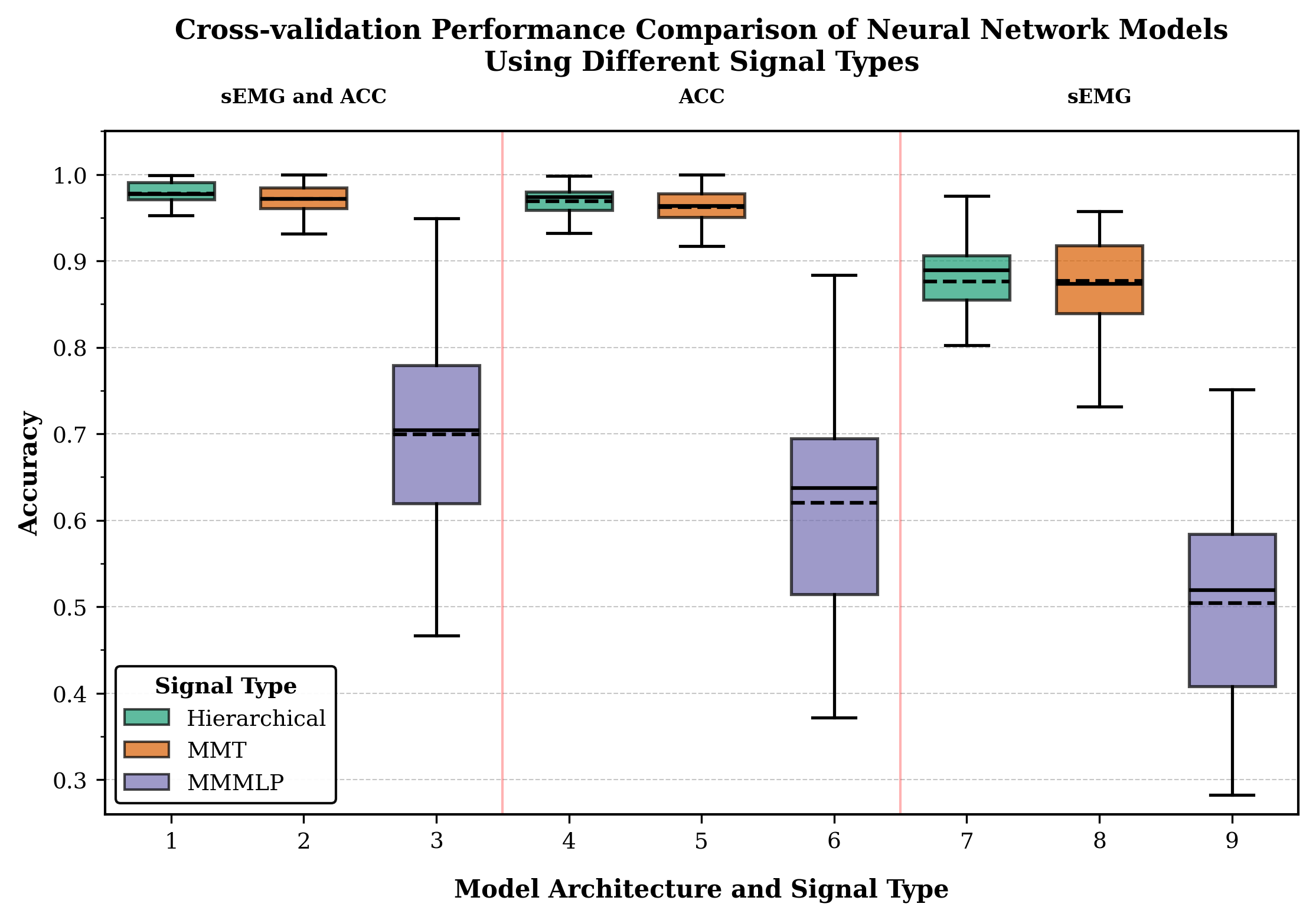}
    \caption{NinaPro~DB2 (sEMG + ACC)}
    \label{fig:EMG+ACC_models}
\end{subfigure}\hfill
\begin{subfigure}[t]{0.48\linewidth}
    \centering
    \includegraphics[width=\linewidth]{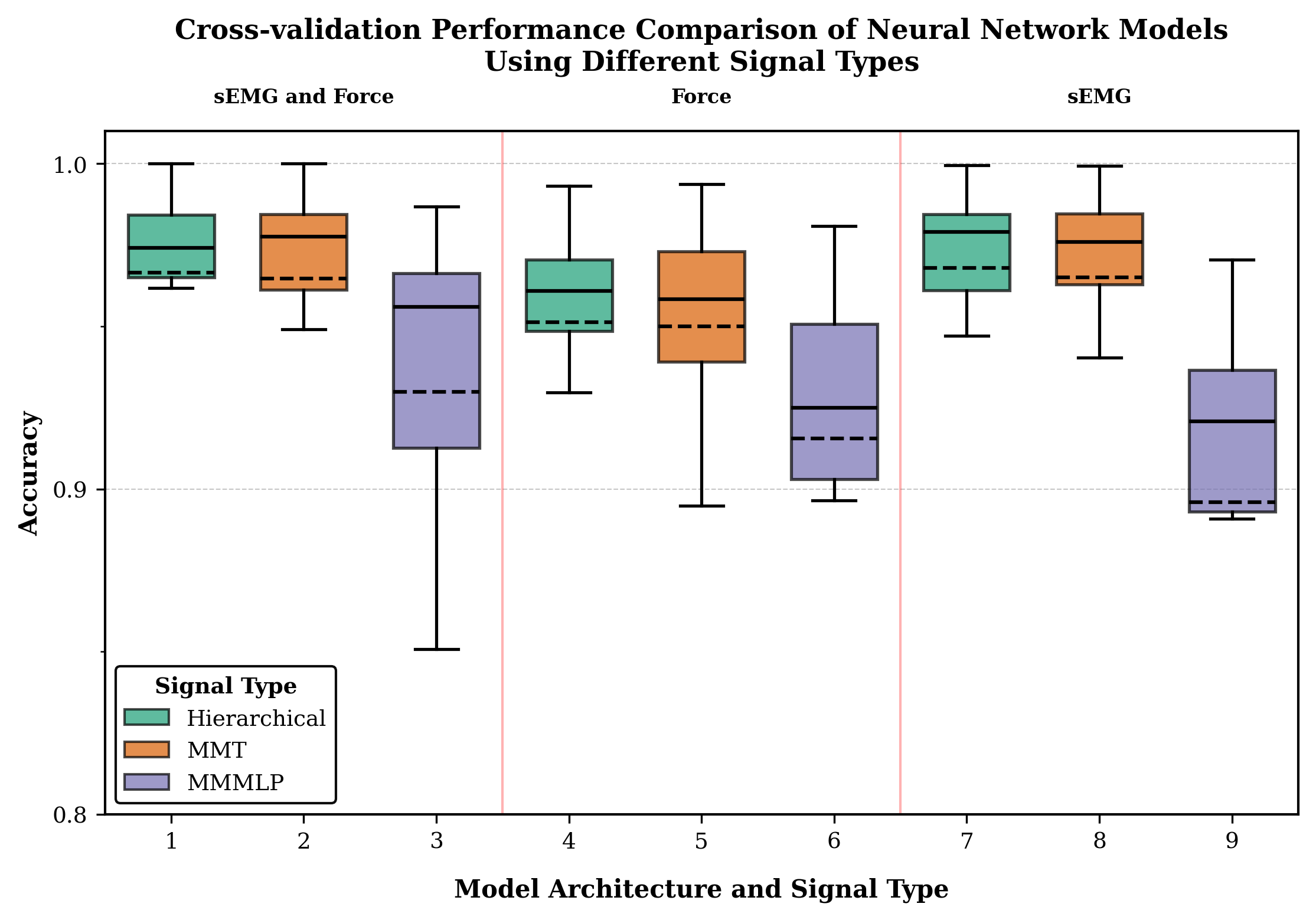}
    \caption{HD-sEMG 65-Gesture (HD-sEMG + Force)}
    \label{fig:EMG+FMG_models}
    \vspace{-0.25cm}
\end{subfigure}

\caption{Accuracy of the Hierarchical Transformer, Multimodal Transformer, and Multimodal MLP on two multimodal datasets: (a) NinaPro DB2 with sEMG + accelerometer inputs and (b) HD-sEMG 65-Gesture with HD-sEMG + force inputs.  Each box plot shows the distribution across all subjects.}
\label{fig:model_comparison_two_datasets}
\end{figure}

Shown in  Fig. \ref{fig:EMG+ACC_models}, the Hierarchical Transformer had the highest overall accuracy from the multimodal signal (mean = 97.76 \%, IQR = 0.0201). Notably, ACC alone performed nearly as well (mean = 97.19 \%), while sEMG alone lagged behind (mean = 69.92 \%). This large disparity is largely due to the ACC modality having 36 channels, three times more than sEMG’s 12, providing richer input. In a controlled 12-channel vs. 12-channel setting (shown in the previous experiment), sEMG actually marginally outperformed ACC, suggesting sEMG's lower performance here is due to limited input dimensionality, not signal quality. The MMT shows similar trends: sEMG and ACC (mean = 96.90 \%) only slightly outperformed ACC alone (mean = 96.21 \%), with sEMG alone again far behind (mean = 61.99 \%). Both transformer models showed consistently high median performance with ACC and multimodal inputs, highlighting the effectiveness of attention mechanisms when applied per modality. The MMMLP\textbf{.} performed significantly worse across the board. sEMG and ACC (mean = 87.60 \%) was nearly identical to ACC alone (mean = 87.66 \%), and sEMG alone dropped to 50.43\%. Variability was higher overall, suggesting that simple MLPs struggle to capture meaningful cross-modal interactions.

\begin{table}[t]
\centering
\captionsetup{justification=centering}
\caption{Statistical tests (a,c) and mean accuracies (b,d) for two multimodal datasets.}
\label{tab:stats_means_two}

\begin{subtable}[t]{0.48\linewidth}
\centering
\captionsetup{justification=centering}
\begin{subtable}[t]{\linewidth}
\caption{Mann--Whitney $U$ test with Bonferroni correction \\ (sEMG + ACC)}
\label{tab:mwu_emg_acc}
\vspace{4pt}
\centering
\begin{tabular}{@{}lll@{}}
\toprule
\textbf{Modality} & \textbf{Comparison} & \textbf{Sig.}\\
\midrule
\multirow{3}{*}{sEMG+ACC}
 & Hier.\ vs.\ MMT   & ns   \\
 & Hier.\ vs.\ MMMLP & ***  \\
 & MMT  vs.\ MMMLP   & **** \\[2pt]
\multirow{3}{*}{ACC}
 & Hier.\ vs.\ MMT   & ns   \\
 & Hier.\ vs.\ MMMLP & ***  \\
 & MMT  vs.\ MMMLP   & ***  \\[2pt]
\multirow{3}{*}{sEMG}
 & Hier.\ vs.\ MMT   & ns   \\
 & Hier.\ vs.\ MMMLP & **** \\
 & MMT  vs.\ MMMLP   & ***  \\
\bottomrule
\end{tabular}
\end{subtable}

\vspace{6pt}
\begin{subtable}[t]{\linewidth}
\caption{Mean accuracies ($N{=}40$)}
\label{tab:mean_emg_acc}
\vspace{4pt}
\centering
\begin{tabular}{@{}lccc@{}}
\toprule
\textbf{Modality} & \textbf{Hier.} & \textbf{MMT} & \textbf{MMMLP}\\
\midrule
sEMG+ACC & 0.9776 & 0.9690 & 0.8760\\
ACC      & 0.9719 & 0.9621 & 0.8766\\
sEMG     & 0.6992 & 0.6199 & 0.5043\\
\bottomrule
\end{tabular}
\end{subtable}
\end{subtable}
\hfill
\begin{subtable}[t]{0.48\linewidth}
\centering
\captionsetup{justification=centering}
\begin{subtable}[t]{\linewidth}
\caption{Mann--Whitney $U$ test with Bonferroni correction \\ (sEMG + Force)}
\label{tab:mwu_emg_force}
\vspace{4pt}
\centering
\begin{tabular}{@{}lll@{}}
\toprule
\textbf{Modality} & \textbf{Comparison} & \textbf{Sig.}\\
\midrule
\multirow{3}{*}{sEMG+Force}
 & Hier.\ vs.\ MMT   & ns \\
 & Hier.\ vs.\ MMMLP & ** \\
 & MMT  vs.\ MMMLP   & *  \\[2pt]
\multirow{3}{*}{Force}
 & Hier.\ vs.\ MMT   & ns \\
 & Hier.\ vs.\ MMMLP & ns \\
 & MMT  vs.\ MMMLP   & ns \\[2pt]
\multirow{3}{*}{sEMG}
 & Hier.\ vs.\ MMT   & ns   \\
 & Hier.\ vs.\ MMMLP & **** \\
 & MMT  vs.\ MMMLP   & ***  \\
\bottomrule
\end{tabular}
\end{subtable}

\vspace{6pt}
\begin{subtable}[t]{\linewidth}
\caption{Mean accuracies ($N{=}20$)}
\label{tab:mean_emg_force}
\vspace{4pt}
\centering
\begin{tabular}{@{}lccc@{}}
\toprule
\textbf{Modality} & \textbf{Hier.} & \textbf{MMT} & \textbf{MMMLP}\\
\midrule
sEMG+Force & 0.9666 & 0.9647 & 0.9298\\
Force      & 0.9514 & 0.9500 & 0.9156\\
sEMG       & 0.9680 & 0.9651 & 0.8960\\
\bottomrule
\end{tabular}
\end{subtable}
\end{subtable}
\vspace{-0.35cm}
\end{table}

For HD-sEMG 65-Gesture we repeated the 3-way comparison with HD-sEMG alone (128 channels), force alone (9 gauges), and their combination, training each model for 500 epochs.
Results appear in Fig. \ref{fig:EMG+FMG_models} and Table \ref{tab:mean_emg_force}.

The Hierarchical Transformer had the best overall performance, though different from NinaPro. Using both modalities it reached 96.66\% accuracy, essentially tied with HD-sEMG alone at 96.80\% (Mann–Whitney: \textit{ns}) and $\approx 1.5\% $above force alone (95.14\%). The negligible gap between multimodal and HD-sEMG-only inputs suggests that the dense 8 × 8 electrode grids already capture most task-relevant variation, leaving limited head-room for the additional force signal. The MMT shows the same pattern: 96.47\% for the fused input versus 96.51\% for HD-sEMG alone (\textit{ns}), again indicating information saturation. Force alone lags behind at 95.00\%, confirming that mechanical output is less discriminative than the high-resolution EMG field. The MMMLP benefits most from fusion: its accuracy jumps from 89.60\% (HD-sEMG) and 91.56\% (force) to 92.98\% when both are present (p < 0.01 versus HD-sEMG; Table \ref{tab:mwu_emg_force}).
Because the MLP cannot exploit spatial patterns in the HD grid, the complementary force signal supplies the missing global context.

Overall, on HD-sEMG the transformers extract nearly all useful information from the EMG grid alone, so adding force yields only marginal gains, whereas the linear model needs the second modality to close part of the accuracy gap.
This contrast with NinaPro—where the lower-density EMG array benefits greatly from an inertial complement—highlights that the value of additional sensors depends on how much unexploited information remains in the primary modality.

To further understand the individual contributions of each modality in the jointly trained Hierarchical Transformer model, we conducted a modality evaluation experiment by zeroing out one input modality at inference time while keeping the model weights fixed. This approach reveals the dependency of each signal stream (e.g., sEMG, ACC, Force) on classification performance. 

\begin{figure}[h]
    \centering
    \includegraphics[width=0.5\linewidth]{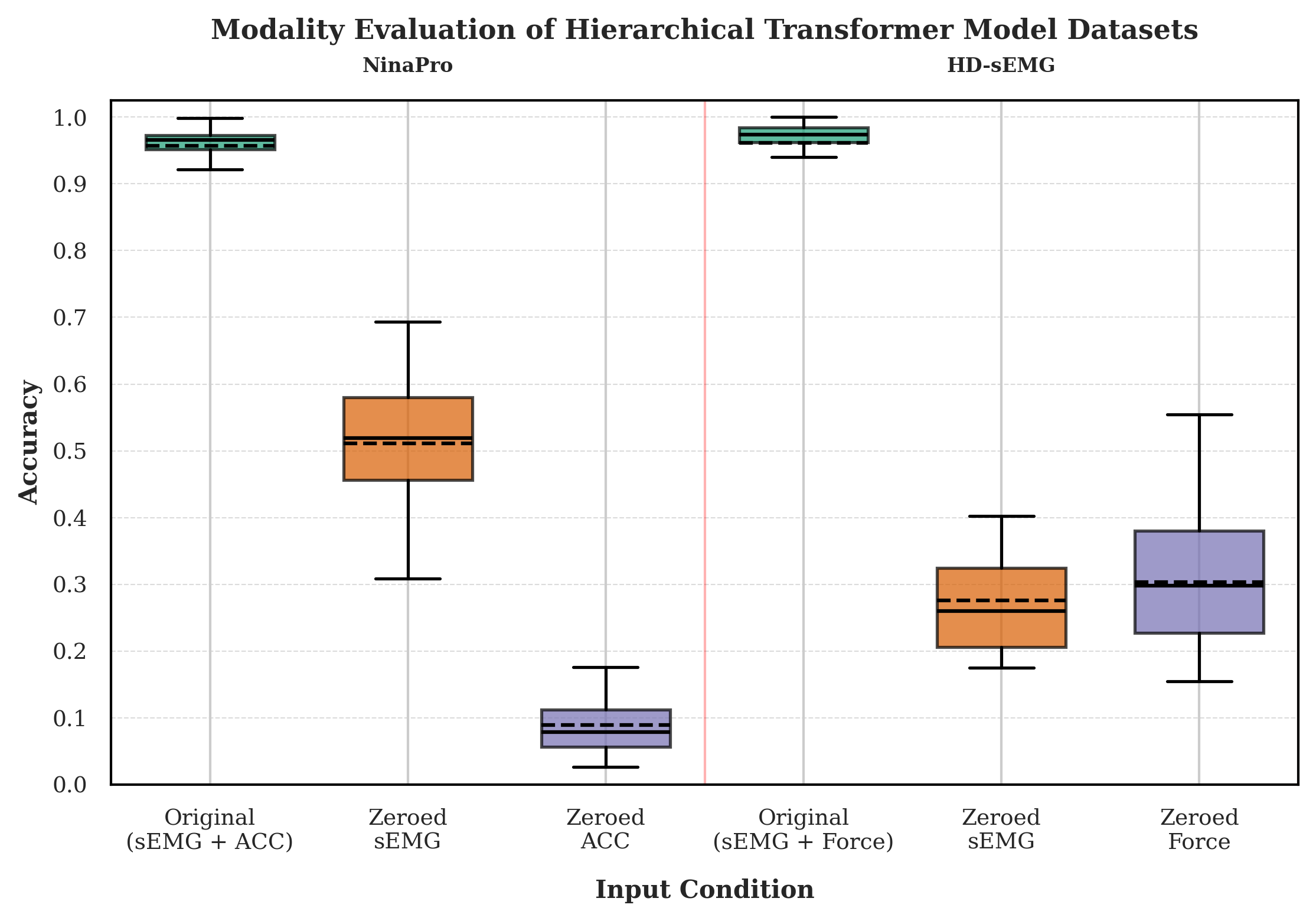}
    \caption{Modality evaluation of the Hierarchical Transformer model on the NinaPro DB2 and HD-sEMG 65 Gesture datasets. The boxplots compare classification accuracy across all subjects in their respective datasets under three input conditions: original (no change to modality input), zeroed modality one (ACC or Force only), and zeroed modality two (sEMG only).}
    \label{fig:Combined_Modal_Eval}
\end{figure}

The results of the modality evaluation experiment on the NinaPro DB2 dataset show that zeroing ACC (sEMG-only condition) leads to a much greater performance drop than zeroing sEMG (ACC only condition), suggesting that ACC provides stronger discriminative features than sEMG. However, both input conditions perform significantly worse than the original sEMG and ACC input. 

In the modality evaluation using the HD-sEMG 65 Gesture dataset, the performance degradation from zeroing either HD-sEMG or Force input was more balanced. Although the Force only condition had slightly higher mean accuracy, it also showed greater variability between subjects. These results suggest that, unlike NinaPro, both modalities contribute comparably to performance, and neither is clearly dominant. The results of the modality evaluation can be seen in Fig. \ref{fig:Combined_Modal_Eval}, with the average accuracies of both datasets under various input conditions summarized in Table \ref{tab:modality-eval}.

\begin{table}[ht]
    \centering
    \caption{Average accuracy for zeroed input evaluations}
    \vspace{4pt}
    \label{tab:modality-eval}
    \begin{tabular}{l l c}
        \toprule
        \textbf{Dataset}         & \textbf{Input Condition}         & \textbf{Avg. Accuracy} \\
        \midrule
        \multirow{3}{*}{\textbf{NinaPro}} 
            & Zeroed ACC (sEMG only)      & 0.0890 \\
            & Zeroed sEMG (ACC only)      & 0.5110 \\
            & Original (sEMG + ACC)       & 0.9571 \\
        \midrule
        \multirow{3}{*}{\textbf{HD-sEMG}} 
            & Zeroed Force (sEMG only)    & 0.3031 \\
            & Zeroed sEMG (Force only)    & 0.2757 \\
            & Original (sEMG + Force)     & 0.9611 \\
        \bottomrule
    \end{tabular}
    
\end{table}

To investigate the role of unimodal and cross-modal attention mechanisms in multimodal gesture classification, we conducted a series of attention masking ablation experiments using the IsoNet architecture on Ninapro DB2 Exercise B gestures. These experiments aimed to isolate how the model's performance is affected by selectively disabling attention flows between and within modalities across different transformer layers. Specifically, we evaluated three masking schemes: individual layer masking, which disables attention only at a single transformer layer; range from beginning (RFB), which progressively masks from the first layer up to the current layer; and range from end (RFE), which masks from the final layer backward to the current one. 

By comparing performance across these masking modes, we sought to answer several key questions: (i) which layers contribute most critically to unimodal and cross-modal token integration; (ii) whether early-layer attentions are foundational or if the model primarily depends on late-layer fusion; and (iii) whether attention is distributed across layers or concentrated at specific bottlenecks. The goal of this analysis is to uncover whether attention-driven feature integration occurs primarily in early or late layers and to evaluate whether token interactions exhibit redundancy, specialization, or a dependency hierarchy over depth.

\begin{figure}[h]
\centering
\begin{subfigure}[b]{0.32\linewidth}
    \includegraphics[width=\linewidth]{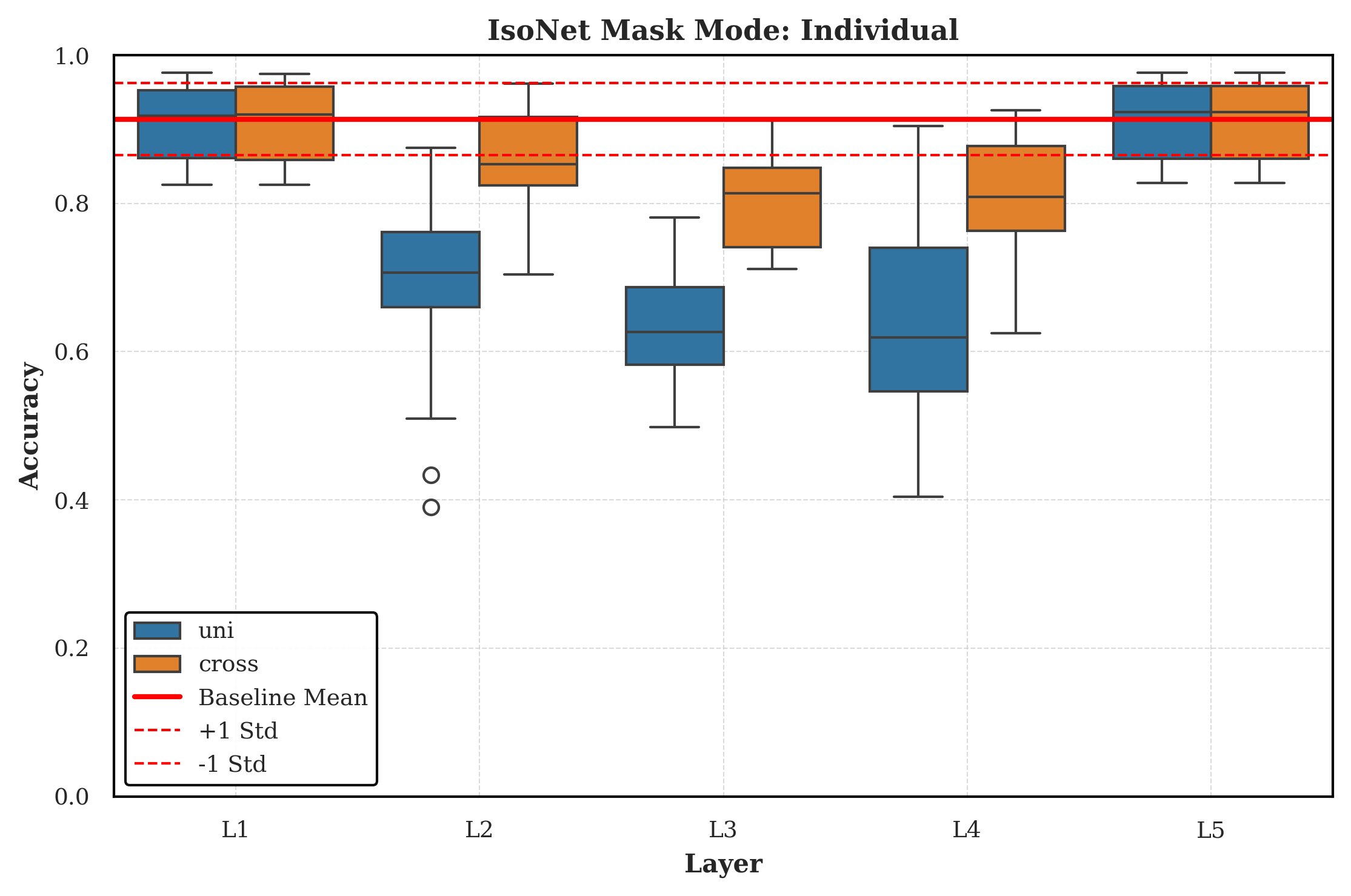}
    \caption{Individual layer masking}
    \label{fig:Individual_Layer_Masking}
\end{subfigure}\hfill
\begin{subfigure}[b]{0.32\linewidth}
    \includegraphics[width=\linewidth]{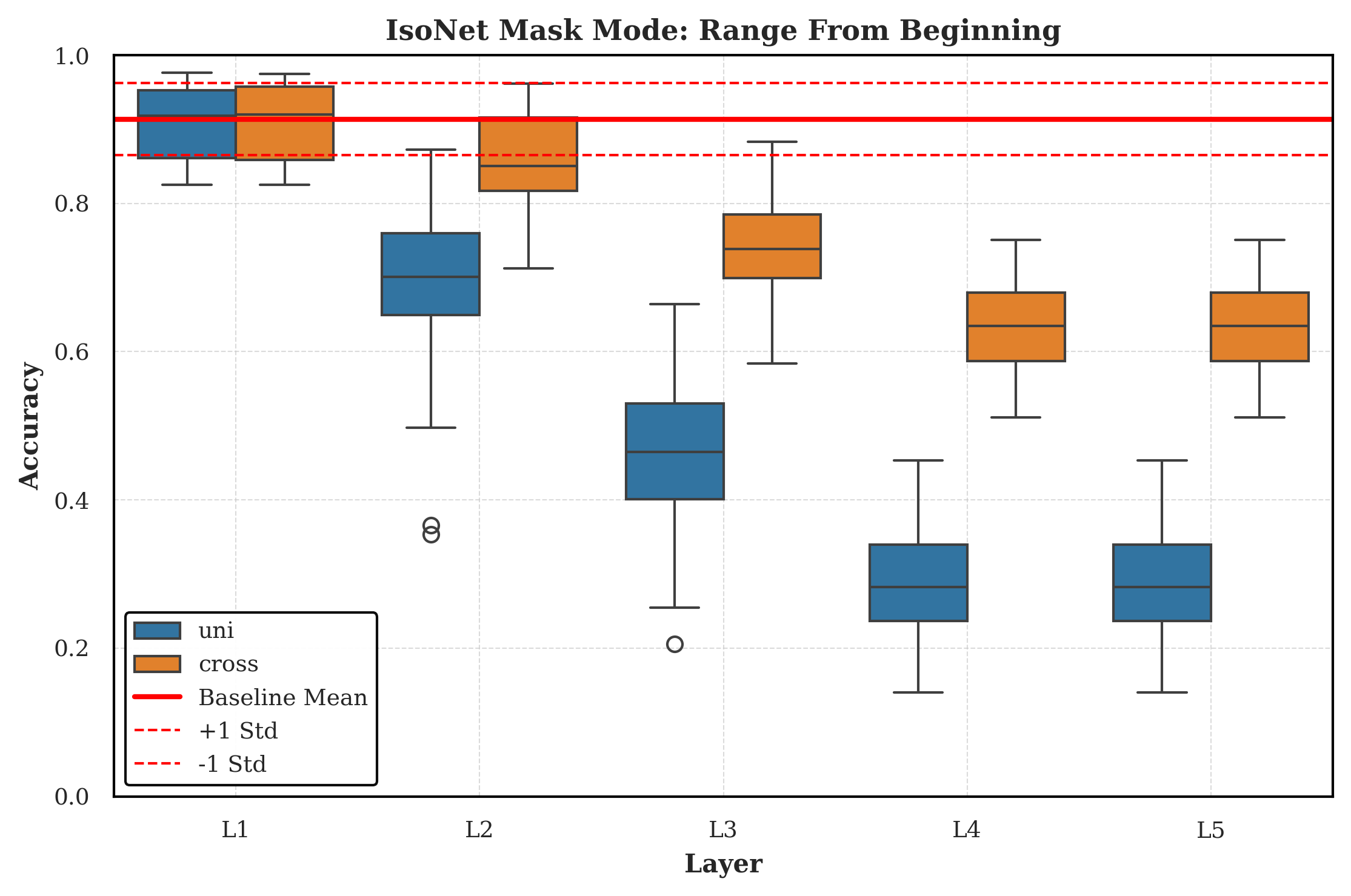}
    \caption{RFB: mask from first layer on}
    \label{fig:RFB_Masking}
\end{subfigure}\hfill
\begin{subfigure}[b]{0.32\linewidth}
    \includegraphics[width=\linewidth]{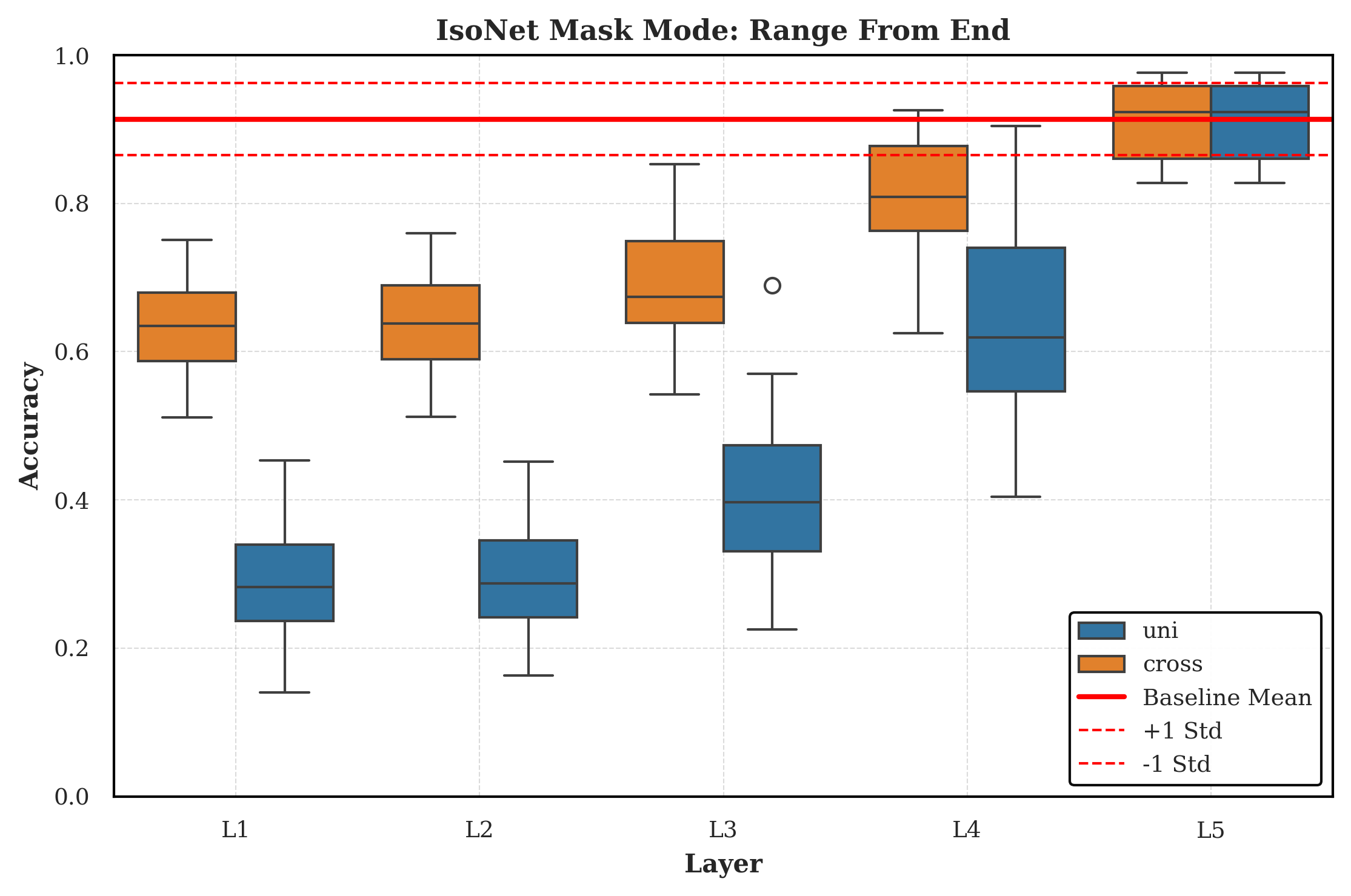}
    \caption{RFE: mask from last layer back}
    \label{fig:RFE_Masking}
\end{subfigure}

\caption{Accuracy boxplots for IsoNet on NinaPro DB2 under three
attention-masking protocols: (a) single-layer masks, (b) cumulative masks from
the start (\textsc{RFB}), and (c) cumulative masks from the end
(\textsc{RFE}).  Unimodal and cross-modal masks appear side-by-side in each
subplot, revealing that mid-depth cross-modal heads cause the largest
accuracy loss.}
\end{figure}

\begin{table}[h]
    \centering
    \caption{Mean accuracy, significance, and percentage drop (\(\Delta^{\%}\)) for attention-masking ablations}
    
    \vspace{4pt}
    \label{tab:masking_ablation}
    \begin{tabular}{lcccccccccc}
        \toprule
        & & \multicolumn{3}{c}{\textbf{Individual}} & \multicolumn{3}{c}{\textbf{RFB}} & \multicolumn{3}{c}{\textbf{RFE}} \\
        \cmidrule(lr){3-5} \cmidrule(lr){6-8} \cmidrule(lr){9-11}
        \textbf{Layer} & \textbf{Type} & \textbf{Acc.} & \textbf{Sig.} & \textbf{$\Delta^{\%}$} 
        & \textbf{Acc.} & \textbf{Sig.} & \textbf{$\Delta^{\%}$} 
        & \textbf{Acc.} & \textbf{Sig.} & \textbf{$\Delta^{\%}$} \\
        \midrule
        \multirow{2}{*}{L1} 
            & Uni & 0.909 & ns & 0.6 & 0.909 & ns & 0.6 & 0.286 & **** & 68.7 \\
        & Cross & 0.911 & ns & 0.3 & 0.911 & ns & 0.3 & 0.636 & **** & 30.4 \\
        \cmidrule(lr){1-11}
        \multirow{2}{*}{L2} 
            & Uni & 0.689 & **** & 24.6 & 0.682 & **** & 25.3 & 0.289 & **** & 68.4 \\
        & Cross & 0.860 & *** & 5.9 & 0.858 & *** & 6.1 & 0.639 & **** & 30.1 \\
        \cmidrule(lr){1-11}
        \multirow{2}{*}{L3} 
            & Uni & 0.637 & **** & 30.3 & 0.460 & **** & 49.7 & 0.406 & **** & 55.6 \\
        & Cross & 0.803 & **** & 12.1 & 0.745 & **** & 18.5 & 0.694 & **** & 24.0 \\
        \cmidrule(lr){1-11}
        \multirow{2}{*}{L4} 
            & Uni & 0.633 & **** & 30.7 & 0.286 & **** & 68.7 & 0.633 & **** & 30.7 \\
        & Cross & 0.813 & **** & 11.1 & 0.636 & **** & 30.4 & 0.813 & **** & 11.1 \\
        \cmidrule(lr){1-11}
        \multirow{2}{*}{L5} 
            & Uni & 0.914 & ns & 0.0 & 0.286 & **** & 68.7 & 0.914 & ns & 0.0 \\
        & Cross & 0.914 & ns & 0.0 & 0.636 & **** & 30.4 & 0.914 & ns & 0.0 \\
        \midrule
        Baseline & -- & 0.914 & -- & -- & -- & -- & -- & -- & -- & -- \\
        \bottomrule
    \end{tabular}
    
\end{table}

The results demonstrate that IsoNet’s attention mechanisms are not uniformly important across depth, with layers exhibiting distinct contributions to unimodal and cross-modal attention. (i) In the individual layer masking experiments in Fig. \ref{fig:Individual_Layer_Masking}, suppressing attention at L1 causes a non-significant drop, while masking L5 leaves accuracy unchanged. This indicates redundancy at the extremes: early features can be rebuilt deeper in the network, and the final layer mainly refines an already separable representation. In contrast, masking L2–L4 produces substantial drops, especially for unimodal edges, identifying these layers as the primary fusion zone. (ii) Comparing RFB and RFE masking, seen in Figs. \ref{fig:RFB_Masking} and \ref{fig:RFE_Masking} respectively, we observe that performance degrades faster in the RFE setting than in RFB, confirming that late-layer attention is more critical than early-layer attention for effective feature fusion. This asymmetry is also evidenced by the significance evaluation between each mask mode and mask type combination compared to the baseline performance seen in Table \ref{tab:masking_ablation}. Furthermore, masking all cross-modal heads in every layer reduces the mean accuracy to 0.636 (\(-30.4\,\%\) of the baseline), showing that cross-modal interactions contribute nearly a third of the predictive signal. (iii) Accuracy generally declines as progressively deeper layers are masked, except for a small uptick when cross-modal edges are removed at L4 (0.813 vs. 0.803 at L3). These findings indicate that IsoNet's attention-driven integration becomes increasingly specialized in deeper layers up to the penultimate layer. Although unimodal attention mechanisms appear to be more dominant than cross-modal ones across the ablation settings, both are necessary to obtain optimal performance.
\vspace{-0.65cm}

\section{Conclusion}

This study demonstrates that even when trained on equal amounts of entirely unique data, a multimodal MLP combining sEMG and accelerometer channels outperforms its unimodal counterparts by 11.1\%, confirming that complementary modalities yield richer features for gesture decoding. Evaluated on NinaPro DB2 and HD-sEMG 65-Gesture, our Hierarchical Transformer then exceeded a matched linear baseline by 10.2\% and 3.68\%, respectively, showing that late-stage self-attention integrates electrical and mechanical cues more effectively than simple concatenation. To dissect these gains, we introduced IsoNet, whose channel-wise embeddings, dual-head annealed loss, and modality-edge masking enable clean causal interventions. We found that 30.4\% of the predictive signal flows through cross-modal heads concentrated in the network’s middle layers with minimal contribution from its first and last layers. The IsoNet results depend on the dataset and serve as baselines for future cross-dataset comparisons.

For better neurosignal processing of muscle activities, we recommend the adoption of a lean, heterogeneous sensor design with fewer components overall, but combining electrical and mechanical modalities to maximally exploit cross-modal interactions. Complementing this hardware insight, our masking intervention defines a new token-group contribution metric. By ranking attention heads by uni- versus cross-modal importance, one can steer training to specialize heads in inter- or intra-modality patterns, an implicit mixture of experts per layer, and thus enhance sensor signal selectivity and specificity under resource constrained settings.

Ongoing work will extend causal masking to feed-forward activations, test subject-independent generalization, and profile latency–energy trade-offs for IsoNet and the Hierarchical Transformer on embedded platforms. Together, we deliver three contributions: attention-driven accuracy gains, depth-resolved cross-modal attribution, and a token-group contribution metric, that pave the way for transparent, resource-aware decoders in neurorobotic control, advanced prosthetic, and mixed-reality systems.

\begin{ack}
The authors would like to acknowledge the help of Dr. S. Farokh Atashzar in the project. Additionally, we have received support from HPC facilities at NYU Tandon.

\end{ack}

\bibliographystyle{unsrtnat}
\bibliography{ref}

\end{document}